%% file: conference_101719.tex
\documentclass[conference]{IEEEtran}
\IEEEoverridecommandlockouts
\usepackage{cite}
\usepackage{amsmath,amssymb,amsfonts}
\usepackage{algorithmic}
\usepackage{graphicx}
\usepackage{textcomp}
\usepackage{xcolor}
\usepackage{cite}
\usepackage{subfigure}
\usepackage[ruled, linesnumbered]{algorithm2e}

\usepackage{soul}
\usepackage{booktabs} 
\usepackage{graphicx}
\usepackage{amsthm}
\usepackage{amsmath}
\usepackage{multirow}
\usepackage{enumerate}
\usepackage{amsfonts}
\usepackage{color}
\usepackage{url}
\usepackage{amsfonts}
\usepackage{tabularx}
\usepackage{diagbox}
\usepackage{longtable}
\usepackage{rotating}
\usepackage{slashbox}
\usepackage[justification=centering]{caption}
\usepackage[ruled, linesnumbered]{algorithm2e}
\usepackage{bm}
\usepackage{float}

\usepackage{hyperref}

\def\BibTeX{{\rm B\kern-.05em{\sc i\kern-.025em b}\kern-.08em
    T\kern-.1667em\lower.7ex\hbox{E}\kern-.125emX}}
\begin{document}

\title{Self-optimizing Feature Generation via Categorical Hashing Representation and Hierarchical Reinforcement Crossing\\
}

\author{\IEEEauthorblockN{Wangyang Ying$^1$\IEEEauthorrefmark{1}, Dongjie Wang$^2$\IEEEauthorrefmark{1}, Kunpeng Liu$^3$, Leilei Sun$^4$, Yanjie Fu$^1$\IEEEauthorrefmark{2}}
\IEEEauthorblockA{\textit{$^1$ School of Computing and Augmented Intelligence, Arizona State University, Tempe, USA} \\
\textit{$^2$ Department of Computer Science, University of Central Florida, Orlando, USA}\\
\textit{$^3$ Department of Computer Science, Portland State University, Portland, USA} \\
\textit{$^4$ Department of Computer Science, Beihang University, Beijing, China} \\
\{wangyang.ying, yanjie.fu\}@asu.edu, dongjie.wang@ucf.edu, kunpeng@pdx.edu, leileisun@buaa.edu.cn}
}

\maketitle
\footnotetext[1]{Both author contributed equally to this research.}
\footnotetext[2]{Corresponding author}
\input{abstract}
\input{keywords}
\input{introduction}
\input{problem_statement}
\input{methodology}
\input{experiment}
\input{related_work}
\input{conclusion}
\bibliographystyle{IEEEtran}
\bibliography{ref}

\end{document}

%% file: abstract.tex
\begin{abstract}
Feature generation aims to generate new and meaningful features to create a discriminative representation space.
A generated feature is meaningful when the generated feature is from a feature pair with inherent feature interaction.  
In the real world, experienced data scientists can identify potentially useful feature-feature interactions, and generate meaningful dimensions from an exponentially large search space, in an optimal crossing form over an optimal generation path. But, machines have limited human-like abilities.
We generalize such learning tasks as self-optimizing feature generation. 
Self-optimizing feature generation imposes several under-addressed challenges on existing systems: meaningful, robust, and efficient generation. 
To tackle these challenges, we propose a principled and generic representation-crossing framework to solve self-optimizing feature generation.
To achieve hashing representation, we propose a three-step approach: feature discretization, feature hashing, and descriptive summarization. 
To achieve reinforcement crossing, we develop a hierarchical reinforcement feature crossing approach.
We present extensive experimental results to demonstrate the effectiveness and efficiency of the proposed method. The code is available at \url{https://github.com/yingwangyang/HRC_feature_cross.git}.
\end{abstract}

%% file: keywords.tex

%% file: introduction.tex
\vspace{-0.3cm}
\section{Introduction}
\vspace{-0.1cm}

Feature generation (FG) is a technique used in machine learning to combine two or more input features in order to create new, more complex features. The new features are created by taking combinations of the original features, such as their concatenation, products, ratios, or differences, and can help capture interactions or nonlinear relationships between the original features. For example, if we have two input features, "age" and "income", we could create a new feature called "age times income" by taking the product of these two variables. This new feature may be more informative than either of the original features alone, as it captures the joint effect of age and income on the outcome variable. Feature generation can be a powerful tool for improving the predictive performance of machine learning models, particularly in cases where the relationships between the input features and the outcome variable are complex or nonlinear. 

There are two major challenges in self-optimizing FG: 1) meaningful and robust generation to improve prediction and fight data bias, 2) automated and efficient generation to mimic human decision behaviors and reduce search space. 
First, except for a given original feature set,  all the principles and mechanisms of FG are unknown: how the optimal target feature set looks like? Which generation path is the best to create a representation space where data patterns are preserved and discriminative? Is the feature-feature crossing form robust enough to fight against outlier or extreme data values?
Under such an open and uncertain environment, meaningful and robust generation seeks to answer: how can we automatically perceive implicit feature interaction,  generate effective new features, and fight against outlier feature values?
Second, the space of FG can exponentially grow. The automated generation suffers from large search space and long calculation time.  Greedy strategies are faster but result in local optimal; exhaustive strategies are globally optimized but unrealistically slow. 
Therefore, automated and efficient generation aims to answer: how can we navigate generation paths with a balance between time costs and global optimum?  

\begin{figure}[bt]
	\includegraphics[width=1.0\linewidth]{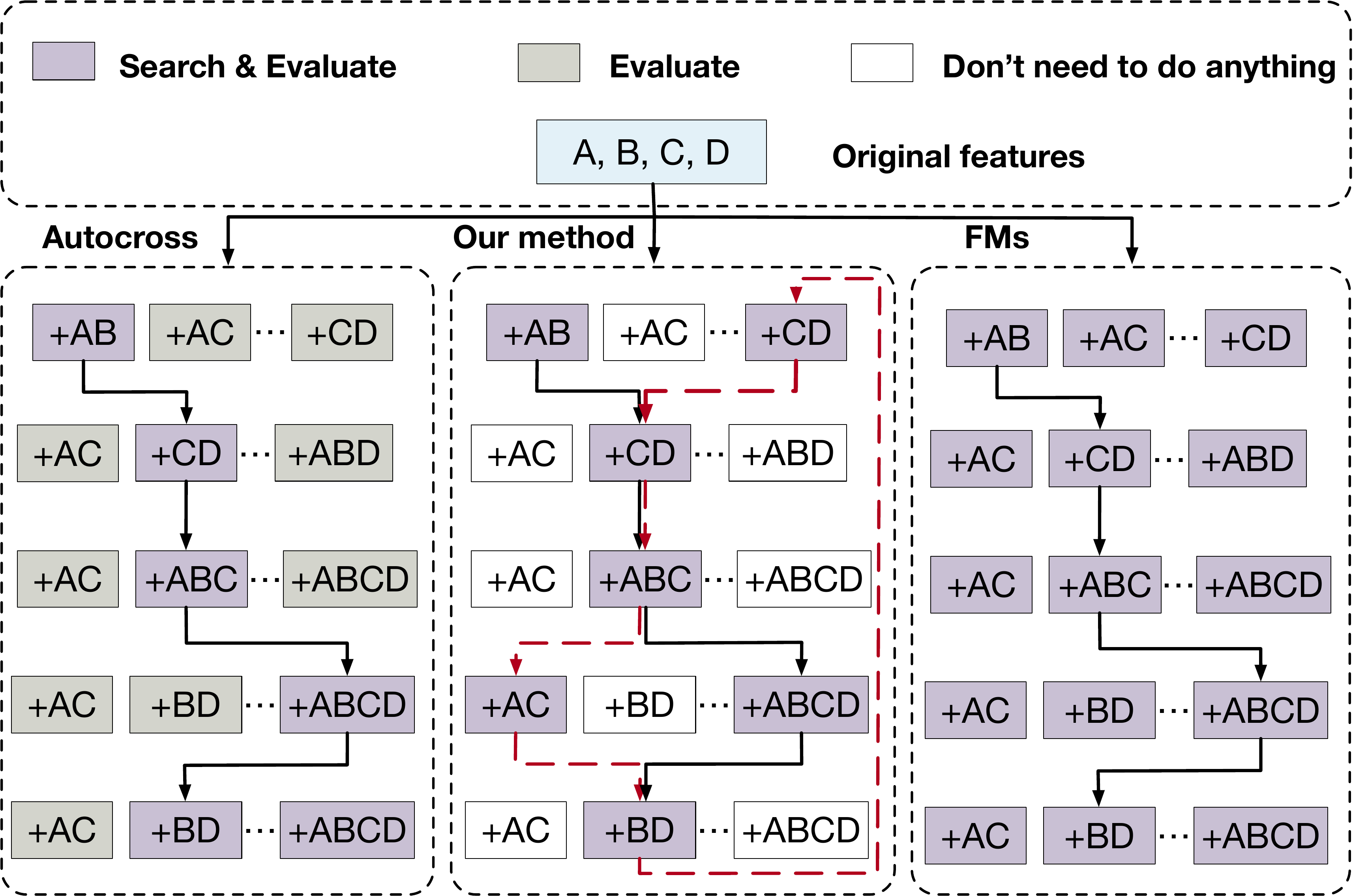}
	\caption{The comparison of search space. Autocross is efficient, but is prone to local optima because it employs a greedy algorithm in each iteration. FMs enumerate all combinations, can find the global optima but are inefficient. Our method does not require exhaustive enumeration as the crossing policies will trim and limit the search space.}
	\label{problem_setting}
 \vspace{-0.5cm}
\end{figure}

As \textbf{Figure \ref{problem_setting}} shows, prior literature can only partially solve the aforementioned challenges.
First, this study is related to exhaustive FG, for example, Factorization Machines (FMs)~\cite{cheng2014gradient,blondel2016higher,juan2016field}, which capture interactions between features from low orders to high orders in order to generate new features.
However, FMs enumerate all feature-feature crossings, and, thus, are inefficient, particularly when a given feature set is large.
Second, this study is related to greedy FG, for instance, AutoCross~\cite{luo2019autocross}, which iterates self-crossing in a previously generated feature set, and always scores and selects the best cross feature based on downstream task performance in each iteration. 
However, while greedy crossing is efficient, its greedy nature exhibits two limitations: 1) only considers the top-1 cross-feature and ignores other cross features; 2) only considers the reward of the current iteration, instead of the long-term reward.
Third, this study is related to reinforcement FG, for instance, GRFG~\cite{wang2022group}, which leverages reinforcement policy to conduct feature-feature crossing. However, existing reinforcement RG is not robust to outlier feature values (e.g., extremely big or small values) when crossing two numeric features. 
Existing studies demonstrate the inability to jointly address meaningful, automated, and efficient generation. 
As a result, we need a new perspective to derive a novel formulation and solver for self-optimizing FG. 

\textbf{Our Contribution: An Integrated Hashing Representation and Reinforcement Crossing Perspective.} 
We observe that a meaningful new feature is usually generated by a feature pair with statistically significant feature-feature interaction. Detection, localization, and measurement of the latent feature interaction are difficult, not to mention designing an explicit crossing form to augment feature space. We show that reinforcement agents are a great fit for sensing the mechanism-unknown feature-feature interaction. 
We formulate FG as a self-optimizing framework to achieve meaningful, robust, and efficient generation. We highlight three contributions in our framework: 1)  hierarchical reinforcement intelligence can learn generation policies to perceive feature-feature interactions, drive the selections of meta feature and crossed feature, and navigate optimal feature generation path, in a way that we reduce search space and balance between global optimal and greedy efficiency.  2)  discretization of feature values before generation is useful because categorical feature crossing can robustilize FG to fight against outlier numeric or continuous feature values to avoid generating anomaly features and introducing bias to future feature generation. 3) the three steps: feature categorization, feature hashing, and descriptive summarization are an efficient strategy to achieve fixed length state representation of a dynamically varying feature space.

\textbf{Summary of Proposed Approach.} 
Inspired by these findings, this paper develops a principled and generic representation-crossing framework for the FG task by iterating hashing-based categorical feature space state representation and hierarchical reinforcement feature crossing (HRC). 
The framework has two goals: 
1) categorizing, hashing, and describing feature space to achieve outlier feature value robustness and fast state extraction in the representation step; 
2) hierarchical reinforcement crossing of meta feature and crossed feature to generate meaningful dimensions in the crossing step. 
To achieve Goal 1, we propose a three-step approach: feature discretization, feature hashing, and descriptive summarization. In particular, we first discretize all the data table values feature by feature, then hash categorical features into a small feature table, and finally exploit dual (feature-wise and instance-wise) descriptive statistics to extract a fixed-length feature space representation. 
Feature discretization can eliminate extreme and outlier values, which could later participate in feature crossing and introduce bias into future generations. 
Unlike one-hot encoding that creates a large feature table, feature hashing can reduce the size of the data table, and help to complete the next step of descriptive summarization in a short time. 
Descriptive summarization can extract a fixed-length state representation of a large dynamically-varying feature space, because DQN usually only takes fixed-length state representation as inputs.  
To achieve Goal 2, we develop a hierarchical reinforcement feature crossing approach. This approach has two agents: meta controller and controller. 
In each iteration, the meta controller selects a meta feature and the controller selects a feature to cross with the meta feature.  The two DQNs of the meta controller and controller learn policies to sense and select two categorical features to cross. 
We leverage mutual information to assess feature-label relevance and feature-feature redundancy, combined with the accuracy of a downstream task as reward quantification to incentivize policy training.  
Finally, we design extensive experiments to verify the effectiveness and efficiency of the proposed methods.

%% file: problem_statement.tex
\vspace{-0.3cm}
\section{Problem Statement}\label{hierarchical_generative_strategy}
\vspace{-0.1cm}

\noindent\textbf{Cartesian Feature Crossing.}
Cartesian crossing is to use the Cartesian join operation to cross categorical features and generate new features. For example, if the two original features are marriage=\{married, single\} and salary = \{high, low\}, then Cartesian join will generate a new feature with four categorical values:  \{married and high, married and low, single and high, single and low\}. 

\noindent\textbf{Feature Generation and Key Challenge.}
The feature generation problem is to generate new features from original features in order to improve feature space and help the downstream model obtain better predictive performance. 
The key challenge is that the number of potential feature sets is extremely large. 
For example, given $N$ original features, if the highest order of generated feature is $k$, then the overall number of original features together with generated features is $C_N^1 + \sum_{k=2}^{N} C_N^k = 2^N -1$. 
With so many features, the potential combination of them, i.e., the number of potential feature sets is $2^{(2^N -1)}$. 
Aside from the large number of candidate-generated feature sets, we need to find the best feature set from all of the $2^{(2^N -1)}$ potential candidates, which is a time-consuming and computationally intensive process. 

\begin{figure}[bt]
        \centering
	\includegraphics[scale=0.47]{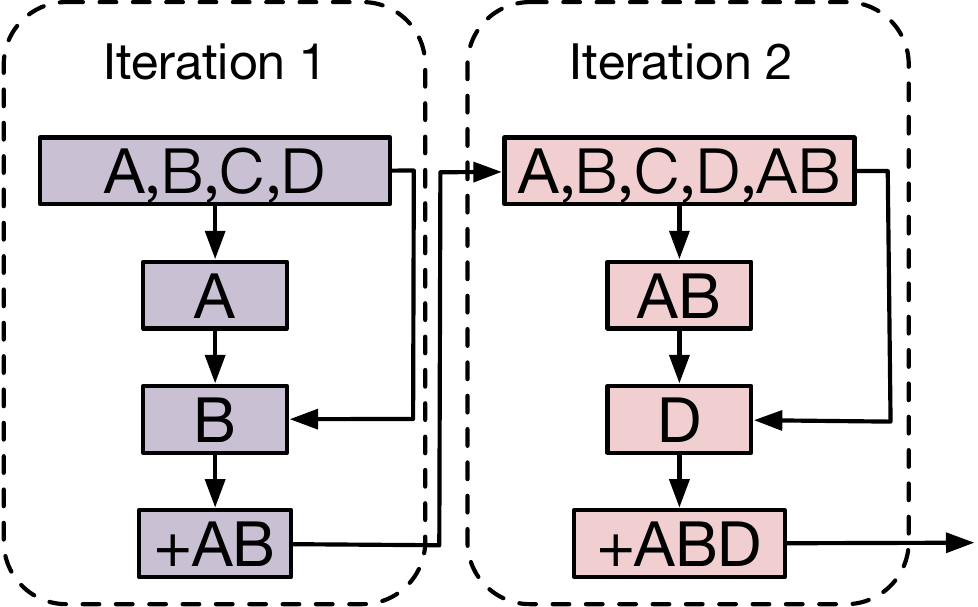}
	\caption{An example of the iterative HRC process.}
	\label{Toy_example}
        \vspace{-0.8cm}
\end{figure}

\noindent\textbf{Iterative Generation to Quickly Approach An Acceptable Optimal.}  Iterative approaches can efficiently generate cross-features to reduce search space and computational costs. 
\textbf{Figure \ref{Toy_example}} shows why iterative generation can allow us to select the most important features and drop the others, and control the number of candidates generated feature sets. 
Specifically, we initialize the optimal generated feature set by the original feature set. In the first iteration, the strategy selects one feature as a meta feature. The strategy then selects another feature that has the strongest interaction with the selected meta feature in order to generate a new feature, which will be added to the generated feature set.
After multiple iterations, we can approach an acceptable optimal generated feature set. 
AutoCross~\cite{luo2019autocross} is also an example study.

\noindent\textbf{Hierarchical Control Structure.} To put this strategy into practice, it's critical to carefully control and manage the selection process. Obviously, there are two crucial stages in each iteration: (1) the initial stage selects a meta feature, (2) and the subsequent stage selects another feature to cross with the meta feature. It is important to note that the two stages are not independent and are not parallel. The first stage needs to smartly sense which meta feature can potentially lead to a new and effective dimension based on the existing feature space; the second stage relies on the first stage and aims to identify another feature that exhibits the strongest interaction with the meta feature.

\noindent\textbf{The Self-optimizing Feature Generation Formulation.}
Formally, given an original feature set and a downstream predictive task, 
we aim to automate the derivation of a generated feature set from the original feature set to maximize the performance improvement of the downstream predictive task. 
We formulate this problem as a hierarchical reinforcement crossing task. 
This framework includes a set of elements: agents, actions, environments, states, and rewards. These elements collaboratively iterate two major steps: 1) feature space representation; 2) crossing policy learning to automate feature generation. Such an interaction-feedback-learning-long-term reward targeting framework is a great fit for solving the joint challenges: 1) the mechanism about how to select two optimal features for crossing is unclear; 2) the generation path needs to trade-off between efficiency and global optimum. 

%% file: methodology.tex
\vspace{-0.1cm}
\section{Self-optimizing Feature Generation}
\vspace{-0.1cm}
\subsection{Framework Overview}
\begin{figure*}[]
	\includegraphics[width=1.0\linewidth]{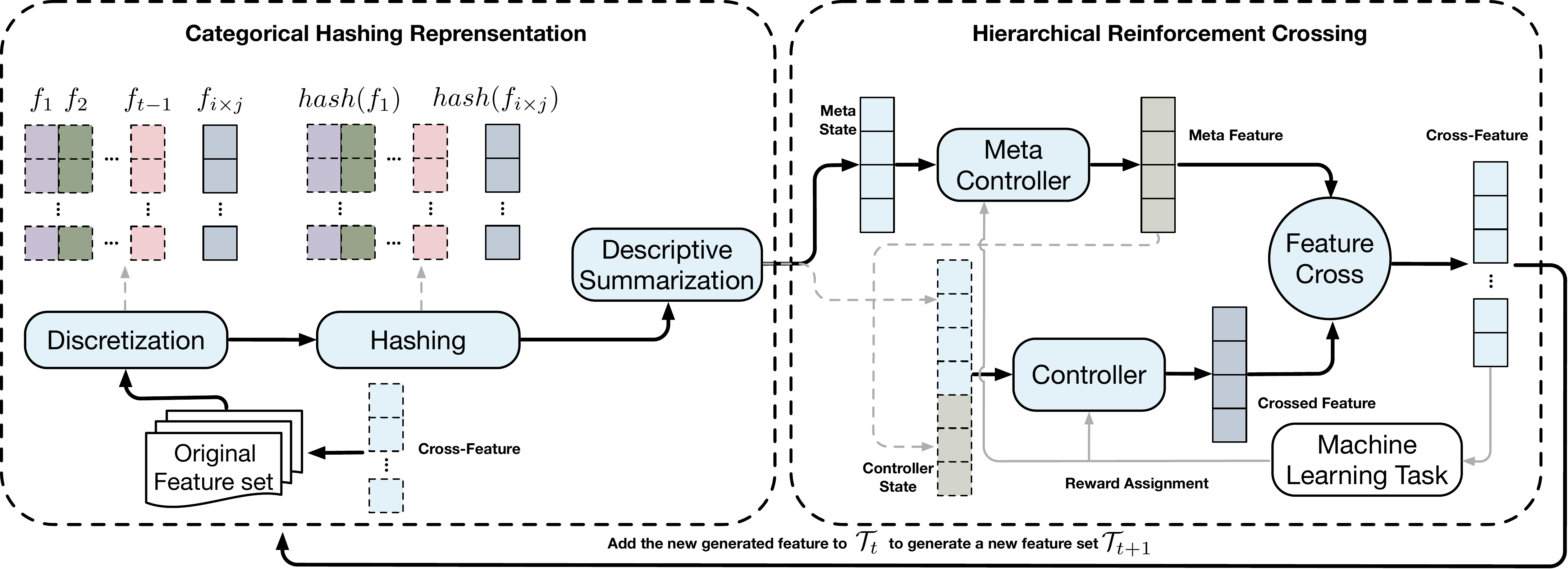}
	\caption{Overview of the framework. In the first part, we convert a dynamically varying feature space into a fixed-length state using the category hashing representation technique. Based on the state, we create a globally optimal sub-feature space using hierarchical reinforcement crossing in the second part.}
	\label{framework}
 \vspace{-0.5cm}
\end{figure*}
\textbf{Figure \ref{framework}} shows our framework iterates two iteratively-interact components: 1) categorical hashing representation, and 2) hierarchical reinforcement crossing. 

The goal of categorical hashing representation is to extract a fixed-length state representation to describe a dynamically varying feature space. 
We propose a three-step approach for feature space state representation by integrating discretization, hashing, and summarization. 
In particular, in Step 1 (feature discretization), we first leverage discretization to convert all the feature values into categorical. This step reformulates FG into pairwise categorical feature crossing that is robust against biased feature generation caused by outlier or extreme continuous feature values crossing.
Since feature values are categorical, if we want to describe the state of feature space, an intuitive solution is to apply one-hot encoding to categorical features in order to ease the statistics extraction of feature space.  However, one-hot encoding can significantly increase the dimensionality of a feature space to describe, making state extraction inefficient.
Therefore, in Step 2 (feature hashing), we propose a faster feature hashing-based approach to encode the categorical feature table into a hashing value table without losing accuracy. 
In Step 3 of descriptive summarization, we integrate both feature-wise and instance-wise descriptive statistics to extract a fixed number of statistics as a fixed-length state representation of the feature space. 
This is because: although the feature space varies over iterations, reinforcement policy learners (e.g., DQN) only accept a fixed-length state as inputs. 

The goal of hierarchical reinforcement crossing is to learn a feature crossing-based operation path to achieve the automated, robust, and fast generation of new meaningful features.
For this purpose, we develop a hierarchical reinforcement crossing approach. 
The reinforcement intelligence is designed for automation: automatically decide how many cross-features to generate in each iteration. 
The hierarchical agent design includes a meta controller agent and a controller agent to sense feature-feature interaction to select a meta feature and a crossed feature that are the most appropriate for crossing new meaningful dimensions. 
The policy-driven generation can improve itself from the feedback of previous generations, find the most optimized direction for future generations, and the tradeoff between greedy search-caused local optimal issues and exhaustive search-caused low-efficiency issues.  

Finally, we iterate both categorical hashing representation and hierarchical reinforcement crossing to conduct self-optimizing feature generation.
\vspace{-0.2cm}
\subsection{Categorical Hashing Representation}\label{state_rep}
\noindent\textbf{Why Categorical Hashing Representation Matters.}
When feature values are continuous data, crossing two continuous features can be easily compromised to generate biased new features. For example, if a feature includes extremely large or small outlier values, these outlier values can propagate bias into generated features. 
We highlight that feature discretization (binning) can aggregate continuous values into category levels, and reduce the bias propagation of extreme values in feature crossing. 
However, if we adopt feature discretization and bin all the data into categorical values, we need to equip reinforcement agents with the ability to describe the state of a categorical feature space. 
Traditional methods, such as neural representation learning or descriptive statistics-based approaches, usually require us to apply one-hot encoding to transform categorical data and ease state representation extraction. 
Unfortunately, one-hot encoding will significantly increase the dimensionality that we will describe as a fixed-length feature space state vector, thereafter, increase the computation burden of state representation extraction. 
So, a dilemma is: how can we fight against outlier and extreme values via discretization, meanwhile improving state representation efficiency? 

\noindent\textbf{Leveraging the Integrated Power of Discretization, Hashing, and Descriptive Summarization.}
We found that integrating feature discretization, feature hashing, and descriptive statistics summarization can achieve both robustness against outlier data, and fast state representation of categorical data. 
Based on our unique insight, we propose a step-by-step testable method
that includes three steps. 
\textbf{Figure \ref{framework}} illustrates the three-step approach of categorical hashing representation.

\noindent\textbf{Step 1: Binning Outliers and Noises.} 
Outliers and noises can introduce and propagate bias in the next and future feature crossing. 
Binning smooths a continuous data value by consulting its neighborhood, transforming the value into a discrete category, and ensuring that data in the same bin is similar and data in different bins are more distinguishable. Therefore, binning can limit noises or anomaly by creating categorical values with distributions comprising fewer unique values, and robustize the crossing of categorical features. 
To this end, we integrate hierarchical bottom-up clustering and $\mathcal{X}^2$-distribution to develop an automated binning approach for feature crossing. Since the feature value distribution after binning should be similar to the feature value distribution before binning,  we leverage a hierarchical bottom-up clustering idea to create small bins first and then combine small bins with similar distributions into large bins. 
Our approach includes two stages: 1) the initialization stage; 2) the bottom-up merging stage.  
In particular, in the initialization stage, we sort the values of a feature according to their numerical values, and then each value of the feature column is treated as a separate bin.
In the merging stage, we iterate the following steps: i) we first calculate the chi-square of each pair of adjacent bins; ii) based on the chi-squares, we merge any neighbor bin pair with the minimum chi-square; iii) we repeat i) and ii) until all the chi-squares of the bins are larger than a certain threshold, or the number of bins reaches a minimum number of bins. 

\noindent\textbf{Step 2: Feature Hashing.}
After Step 1, all the features in the feature space are categorical (nominal). To extract the feature space state representation, we need a numeric representation of a categorical feature table that not only preserves feature space patterns, but also eases the statistical summarization of a feature space. An intuitive way is to use the existing one-hot encoding to map the categorical features into a one-hot data table for statistical summarization without losing too much information. However, one-hot encoding will convert a single categorical feature into multiple features, and dimensionality will increase. Moreover, since iterative feature crossing will keep generating more feature value categories, the dimensionality under one-hot encoding and the time costs of computing the feature space state will explode. 
Our idea is to develop a hashing representation for the categorical features. 
Since a hashing function can map all categorical data into a smaller fixed set, the use of hashing can greatly reduce the number of categories and prevent a dimensional explosion.
Specifically, we use a hash function~\cite{hash} to convert each feature categorical value to an integer value, and then we reduce the integer value modulo a pre-defined number to get a new smaller number.  

\begin{figure*}[]
	\centering
	\includegraphics[width=1.0\textwidth]{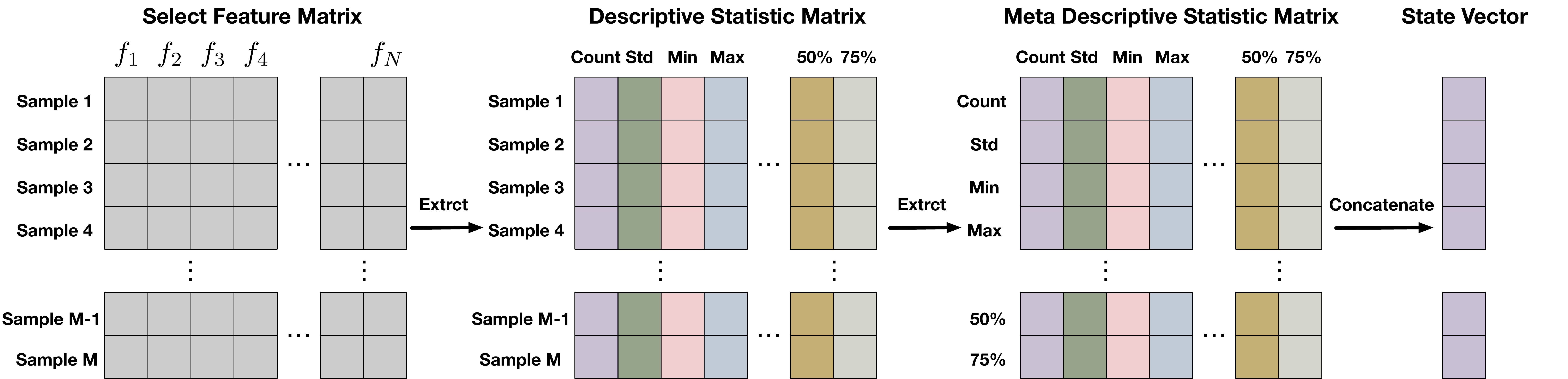}
	\caption{Representation of generated feature set. We extract descriptive statistics twice from the feature set to obtain a fixed-length state vector. The length of the represented state vector remains fixed regardless of the change in feature numbers.}
	\label{fig:Feature_Describe_of_Describe}
 \vspace{-0.5cm}
\end{figure*}

\noindent\textbf{Step 3: Descriptive Summarization.}
 The feature number (dimensionality) of the generated feature set changes at each iteration. However, the policy network and target network in DQN require the feature space stat representation to be a fixed-length vector at each step. 
The goal of Step 3 is to derive a fixed-length state representation of the hashed feature space whose dimension changes over time.
As shown in \textbf{Figure \ref{fig:Feature_Describe_of_Describe}} We propose a descriptive statistics-based dual summarization method that includes two steps: 1) row-wise summarization, and 2) column-wise summarization.
Here, the eight descriptive statistics include:  count, mean, standard, minimum, maximum, and three quartiles (25\%,50\%,75\%).
\textbf{Figure \ref{fig:Feature_Describe_of_Describe}
} shows the two-step procedure. 
Step 1 extracts the eight descriptive statistics of each row in $\mathcal{D}$, and thus, obtains a descriptive statistics matrix with columns as eight statistics and rows as samples. 
Step 2 extracts the eight descriptive statistics of each column of the descriptive statistics matrix, and obtains a meta descriptive statistics matrix with rows as eight statistics and columns as eight statistics. Finally, we concatenate each column of the meta matrix together as the fixed-length state representation vector with a length of 8x8=64.
\vspace{-0.2cm}
\subsection{Hierarchical Reinforcement Crossing}

\noindent\textbf{Why Hierarchical Reinforcement Crossing Matters.}
First, crossing categorical features can generate more explainable and meaningful new dimensions. For instance, a driver age feature with three categorical values (i.e., junior, youth, adult) crosses a driver marriage feature with three categorical values (i.e.,  married, single, divorced), thereafter, generates a new dimension with eight meaningful categorical values  (e.g., married and adult). Such dimension is more meaningful and explainable because it describes a subpopulation (e.g, married and adult drivers) of data with unique driving safety patterns in a feature space. 
Second, the latent interaction between two features is critical to decide whether we should cross them. However, the feature interaction is difficult to measure. Reinforcement is a great tool of AI for decision science when the mechanism is unclear, as it self-learns neural policies from feedback. Besides, a current feature generation step depends on previously generated feature space, which can be viewed as a Markov Decision Process. 
Third, as we need two agents to select two features for crossing, a simplified idea is to assume two agents are independent. Our large-scale analysis shows that the state of feature space changes in a chain, when a feature is selected to cross (we call meta feature), and continues to change when another feature is selected to be crossed (we call crossed feature). The two agents form a hierarchical structure.

\noindent\textbf{Leveraging Categorical Crossing, Self-optimizing Reinforcement, and Hierarchical Agent Structure.}
We propose to develop a hierarchical reinforcement crossing framework that includes a meta controller agent and a controller, each of which in each iteration takes a state $s$ as input, conducts an action $a$, and generates a new feature to obtain an updated generated feature set $\mathcal{T}$. The key components of our proposed method are as follows:

\noindent\underline{\emph{1) State.}} The state $s_t$ is defined as the representation of the generated feature set $\mathcal{T}_t$. The agent makes decisions based on the current generated feature set $\mathcal{T}_t$ to produce a better $\mathcal{T}_{t+1}$. The algorithm of state representation is in  Section \ref{state_rep}.

\noindent \underline{\emph{2) Action.}} \textbf{Figure \ref{framework}} shows the agent consists of two components: a meta controller and a controller. The action of the meta controller is defined as $a_{1,t}=f_i$: select the $i$-th feature as the meta feature from the $\mathcal{T}_t$. The action of the controller is defined as $a_{2,t}=f_j$: combined with the meta feature, the controller visits all features in the $\mathcal{T}_t$ to select the $j$-th feature that can be crossed with meta feature to improve the representation of the feature set.

\noindent\underline{\emph{3) Supervised and Unsupervised Signals as Rewards.}} 
We found that both supervised signals and unsupervised signals can be reward measurements. 

From the unsupervised signal perspective, we propose to measure feature-feature redundancy as a utility of feature space. Specifically, when two features have very high mutual information, that means the two features are similar and overlapped with a high collinearity. Therefore, adding the two similar features into the same feature space will increase information redundancy, instead of new information. In other words, one of the objectives of FG is to minimize feature redundancy. 
Formally, given a feature set $\mathcal{F}$, we can derive its overall mutual information $Rd(\mathcal{F})$, which we call redundancy:
        \begin{small}
	\begin{equation}\label{Rd}
	Rd(\mathcal{F}) = \frac{1}{|\mathcal{F}|^2}\sum_{f_i,f_j\in \mathcal{F}}^{} I(f_i,f_j)
	\end{equation}
        \end{small}
where $f_i$ is the $i$-th feature and $f_j$ is the $j$-th feature,  $I(f_i,f_j)$ is the mutual information to measure the correlation between two random variables $f_i$ and $f_j$.

From the supervised signal perspective, we propose two reward measurements: 1) feature relevancy; 2) downstream task accuracy. 
First, feature relevancy describes how predictable are labels given a feature. The higher the feature relevancy is, the higher utility the feature has. So, one of the objectives is to maximize feature relevancy. 
Formally, given a feature set $\mathcal{F}$ and the label $y$, we can derive its overall mutual information $Rv(\mathcal{F},y)$, which we call relevance:
        \begin{small}
        \begin{equation}\label{Rv}
	Rv(\mathcal{F},y) = \frac{1}{|\mathcal{F}|}\sum_{f_i\in \mathcal{F}}^{} I(f_i,y)
	\end{equation}
        \end{small}
where $f_i$ is the $i$-th feature and $y$ is the label.
Second, downstream task accuracy is clearly a signal that describes the utility of a feature space. Therefore, another objective is to maximize downstream task accuracy. 

\noindent\ul{\emph{4) Reward Functions for the Meta Controller Agent and Controller Agent.}} 
Since the framework has two agents: the meta controller and the controller, we design two personalized reward functions for the meta controller and controller. 
\begin{itemize}
    \item {\emph{The meta controller reward function}} is quantified by a combination of the accuracy $Acc$ on the machine learning task $\mathcal{M}$ at the current iteration, $Acc_{best}$ on the $\mathcal{M}$ at one window time, the relevance between the generated feature set and the target label, and the redundancy of the generated feature set, given by:
	\begin{equation}\label{meta_reward}
	r_1 = w_1 * (Acc - Acc_{best}) + w_2 * Rv - w_3 * Rd
	\end{equation}
	where $w_i$ ($i\in{1,2,3}$) is a positive weight.
	\item{\emph{The controller reward function}} is measured by the accuracy of the machine learning task $\mathcal{M}$, relevancy, and the redundancy: 
	\begin{equation}\label{controller_reward}
	r_2 = w_4 * (Acc - Acc_{best}) + w_5 * Rv - w_6 * Rd
	\end{equation}
	where $w_i$ ($i\in{4,5, 6}$) is a positive weight.
 \end{itemize}

\noindent \textbf{Temporal Abstraction.}
\textbf{Figure \ref{framework}} shows one iteration of HRC. The meta controller receives the state which is the representation of the current generated feature set, and selects a meta feature from the feature set. Then, the controller selects another feature based on the state and meta feature to generate the cross-feature. Finally, add the cross-feature to the feature set and start the next iteration.
The objective of the meta controller is to maximize the cumulative rewards $r_1$: 
\begin{equation}
    O_1 = \sum{_{t^\prime = 0}^{\infty}} \gamma_1^{t^\prime }  r_{1,t^\prime}
\end{equation}
where $t^\prime$ is the time step, and $\gamma_1$ is a discount factor. Similarly, the objective of the controller is to maximize the cumulative rewards $r_2$: 
\begin{equation}
    O_2 = \sum{_{t^{\prime\prime} = 0}^{\infty}} \gamma_2^{t^{\prime\prime}} r_{2,t^{\prime\prime}}
\end{equation}
where $t^{\prime\prime}$ is the time step and $\gamma_2$ is a discount factor. 

\noindent \textbf{Solving the Model Training Problem.}
We adapt the DQN to derive policies for both the meta controller and the controller. The Q function for the meta controller is given by:

\begin{small}
\begin{equation}
\hspace{-2mm}
\begin{aligned}
 Q_1(s_t,a_{1,t}) &= \text{max}_{a_{1,t}} \sum{_{t^\prime = t}^{\infty}} \gamma_1^{t^\prime -t}  r_{1,t^\prime}\\
 		  &=\text{max}_{a_{1,t}} [r_{1,t}  + \sum{_{t^\prime = t+1}^{\infty}} \gamma_1^{t^\prime-(t+1)}  r_{1,t^\prime}]\\
		  &=\text{max}_{a_{1,t}} [ r_{1,t} +  \gamma_1\text{max}_{a_{1,t+1}} Q_1(s_t,a_{1,t+1}) ]
\end{aligned}
\end{equation}
\end{small}
The Q function for the controller is given by:
\begin{small}
\begin{equation}
Q_2(s_t,a_{2,t}) =\text{max}_{a_{2,t}} [ r_{2,t} +  \gamma_2\text{max}_{a_{2,t+1}} Q_2(s_t,a_{2,t+1}) ]
\end{equation}
\end{small}
We derive the loss function of $Q_1$ based the method in~\cite{mnih2015human}:
\begin{small}
\begin{equation}\label{loss_1}
\hspace{-2mm}
\begin{aligned}
	L_1(\theta_{1,t}) =& E_{(s,a,r,s^\prime) \sim \mathcal{D}_1} \\
       & [(r+\gamma_1\text{max}_{a^\prime} Q_1(s^\prime, a^\prime;\theta_{1,t}^{-}) - Q_1(s, a;\theta_{1,t})^2]
\end{aligned}
\end{equation}
\end{small}
where $\theta_{1,t}^{-}$ are fixed parameters of $\theta_{1,i}$ from the previous iteration. Similarly, we derive the loss function of $Q_2$:
\begin{small}
\begin{equation}\label{loss_2}
\begin{aligned}
L_2(\theta_{2,t}) =& E_{(s,a,r,s^\prime) \sim \mathcal{D}_2} \\
& [(r+\gamma_1\text{max}_{a^\prime} Q_1(s^\prime,a^\prime;\theta_{2,t}^{-})-Q_1(s,a;\theta_{2,t})^2]
\end{aligned}
\end{equation}
\end{small}

We design two independent neural networks for $Q_1$ and $Q_2$ and train them alternately. When making decisions, we adapt a $\epsilon$-greedy algorithm to enhance the exploration ability of reinforcement learning algorithms. The decision history of each network is stored in a memory. The new samples come into the memory and flush the old samples to update the memory. When samples are needed in training, we sample a batch of historical data from the memory and use the gradient descendent technique to update weights in the neural network. We also strategically design a memory mechanism to memorize the historically best version of the generated feature set. In this way, we can still guarantee a relatively good generated feature set, even if HRC does not converge.

%% file: experiment.tex
\vspace{-0.1cm}
\section{Experiment}
\vspace{-0.1cm}
\subsection{Experiment Setup}

\subsubsection{Downstream Tasks}
We validate our proposed feature generation method on classification tasks compared with state-of-the-art baselines. Specifically, we realize classification by three algorithms: logistic regression, decision tree, and random forest. All algorithms are implemented by the scikit-learn package~\cite{scikit-learn}.

\subsubsection{Data Description}
For classification tasks, we use the following four publicly available datasets:
\begin{itemize}
    \item {\textbf{Access}\footnote{https://www.kaggle.com/c/amazon-employee-access-challenge/data}: This dataset records the access to resources of Amazon employees from 2010 to 2011. Label 1 denotes allowed access and Label 0 denotes denied access.}
    \item {\textbf{Bank}\footnote{https://www.kaggle.com/brijbhushannanda1979/bank-data}: This dataset is related to the direct marketing campaigns of a Portuguese banking institution. The classification goal is to predict if a client will subscribe to a term deposit.}
    \item {\textbf{Credit}\footnote{https://www.kaggle.com/c/GiveMeSomeCredit/data}: This dataset includes financial activities of bank customers. The classification goal is to predict if the customer will experience financial distress in the next two years.}
    \item {\textbf{Nomao}\footnote{https://www.openml.org/d/1486}: This dataset is a place localization dataset from UCI. The class label is categorical values that range from 1 to 2, which represents two geographical spots.}
\end{itemize}
All the details about the datasets are shown in \textbf{Table \ref{datasets}}.

\setlength{\tabcolsep}{3.5mm}{
\begin{table}[bt]
\caption{Details of datasets used in the experiments}
\centering
\begin{tabular}{|lcccc|}
\hline
\multicolumn{5}{|c|}{Datasets}                                                                                                                                          \\ \hline
\multicolumn{1}{|c|}{\multirow{2}{*}{Name}} & \multicolumn{2}{c|}{\# Samples}                              & \multicolumn{2}{c|}{Features}                              \\ \cline{2-5} 
\multicolumn{1}{|c|}{}                      & \multicolumn{1}{l|}{Training} & \multicolumn{1}{l|}{Testing} & \multicolumn{1}{l|}{\# Num} & \multicolumn{1}{l|}{\# Cate} \\ \hline
\multicolumn{1}{|l|}{Access}                & \multicolumn{1}{c|}{26,216}   & \multicolumn{1}{c|}{6,553}   & \multicolumn{1}{c|}{0}      & 9                            \\ \hline
\multicolumn{1}{|l|}{Bank}                  & \multicolumn{1}{c|}{21967}    & \multicolumn{1}{c|}{5,492}   & \multicolumn{1}{c|}{10}     & 10                           \\ \hline
\multicolumn{1}{|l|}{Credit}                & \multicolumn{1}{c|}{120,000}  & \multicolumn{1}{c|}{30,000}  & \multicolumn{1}{c|}{10}     & 0                            \\ \hline
\multicolumn{1}{|l|}{Nomao}                 & \multicolumn{1}{c|}{27,572}   & \multicolumn{1}{c|}{6,893}   & \multicolumn{1}{c|}{89}     & 30                           \\ \hline
\end{tabular}
\label{datasets}
\vspace{-0.5cm}
\end{table}}

\begin{center}
\setlength{\tabcolsep}{3.0mm}{
\begin{table*}[]
\caption{Performance comparison of Hierarchical Reinforcement Crossing on different tasks evaluated by accuracy.}
\begin{tabular}{|c|cccc|cccc|cccc|}
\hline
\multirow{2}{*}{} & \multicolumn{4}{c|}{Logistic Regression}                                                                                         & \multicolumn{4}{c|}{Decision Tree}                                                                                               & \multicolumn{4}{c|}{Random Forest}                                                                                               \\ \cline{2-13} 
                  & \multicolumn{1}{c|}{Access}         & \multicolumn{1}{c|}{Bank}           & \multicolumn{1}{c|}{Credit}         & Nomao          & \multicolumn{1}{c|}{Access}         & \multicolumn{1}{c|}{Bank}           & \multicolumn{1}{c|}{Credit}         & Nomao          & \multicolumn{1}{c|}{Access}         & \multicolumn{1}{c|}{Bank}           & \multicolumn{1}{c|}{Credit}         & Nomao          \\ \hline
Raw               & \multicolumn{1}{c|}{0.914}          & \multicolumn{1}{c|}{0.888}          & \multicolumn{1}{c|}{0.900}          & 0.896          & \multicolumn{1}{c|}{0.922}          & \multicolumn{1}{c|}{0.896}          & \multicolumn{1}{c|}{0.909}          & 0.910          & \multicolumn{1}{c|}{0.938}          & \multicolumn{1}{c|}{0.901}          & \multicolumn{1}{c|}{0.915}          & 0.924          \\ \hline
DeepFM            & \multicolumn{1}{c|}{0.921}          & \multicolumn{1}{c|}{0.897}          & \multicolumn{1}{c|}{0.906}          & 0.910          & \multicolumn{1}{c|}{0.934}          & \multicolumn{1}{c|}{0.903}          & \multicolumn{1}{c|}{0.918}          & 0.914          & \multicolumn{1}{c|}{0.941}          & \multicolumn{1}{c|}{0.903}          & \multicolumn{1}{c|}{0.923}          & 0.929          \\ \hline
xDeepFM           & \multicolumn{1}{c|}{0.927}          & \multicolumn{1}{c|}{0.901}          & \multicolumn{1}{c|}{0.911}          & 0.909          & \multicolumn{1}{c|}{0.939}          & \multicolumn{1}{c|}{0.906}          & \multicolumn{1}{c|}{0.923}          & 0.920          & \multicolumn{1}{c|}{0.947}          & \multicolumn{1}{c|}{0.904}          & \multicolumn{1}{c|}{0.925}          & 0.931          \\ \hline
AutoCross         & \multicolumn{1}{c|}{0.930}          & \multicolumn{1}{c|}{0.903}          & \multicolumn{1}{c|}{0.919}          & 0.921          & \multicolumn{1}{c|}{0.941}          & \multicolumn{1}{c|}{0.908}          & \multicolumn{1}{c|}{0.931}          & 0.928          & \multicolumn{1}{c|}{0.951}          & \multicolumn{1}{c|}{0.910}          & \multicolumn{1}{c|}{0.929}          & 0.934          \\ \hline
GRFG         & \multicolumn{1}{c|}{0.935}          & \multicolumn{1}{c|}{0.907}          & \multicolumn{1}{c|}{0.918}          & 0.926          & \multicolumn{1}{c|}{0.941}          & \multicolumn{1}{c|}{0.911}          & \multicolumn{1}{c|}{0.929}          & 0.933          & \multicolumn{1}{c|}{0.946}          & \multicolumn{1}{c|}{0.908}          & \multicolumn{1}{c|}{0.927}          & 0.937         \\ \hline
\textbf{HRC}      & \multicolumn{1}{c|}{\textbf{0.943}} & \multicolumn{1}{c|}{\textbf{0.913}} & \multicolumn{1}{c|}{\textbf{0.938}} & \textbf{0.959} & \multicolumn{1}{c|}{\textbf{0.948}} & \multicolumn{1}{c|}{\textbf{0.915}} & \multicolumn{1}{c|}{\textbf{0.937}} & \textbf{0.953} & \multicolumn{1}{c|}{\textbf{0.955}} & \multicolumn{1}{c|}{\textbf{0.919}} & \multicolumn{1}{c|}{\textbf{0.935}} & \textbf{0.969} \\ \hline
\end{tabular}
\label{overall_performance}
\end{table*}}
\end{center}

\begin{center}
\setlength{\tabcolsep}{3.15mm}{
\begin{table*}[]
\caption{Performance comparison of ablation studies on different tasks evaluated by accuracy.}
\begin{tabular}{|c|cccc|cccc|cccc|}
\hline
\multirow{2}{*}{}    & \multicolumn{4}{c|}{Logistic Regression}                                                                                         & \multicolumn{4}{c|}{Decision Tree}                                                                                               & \multicolumn{4}{c|}{Random Forest}                                                                                               \\ \cline{2-13} 
                     & \multicolumn{1}{c|}{Access}         & \multicolumn{1}{c|}{Bank}           & \multicolumn{1}{c|}{Credit}         & Nomao          & \multicolumn{1}{c|}{Access}         & \multicolumn{1}{c|}{Bank}           & \multicolumn{1}{c|}{Credit}         & Nomao          & \multicolumn{1}{c|}{Access}         & \multicolumn{1}{c|}{Bank}           & \multicolumn{1}{c|}{Credit}         & Nomao          \\ \hline
HRC$^*$           & \multicolumn{1}{c|}{0.939}          & \multicolumn{1}{c|}{0.903}          & \multicolumn{1}{c|}{0.934}          & 0.945          & \multicolumn{1}{c|}{0.940}           & \multicolumn{1}{c|}{0.903}          & \multicolumn{1}{c|}{0.931}          & 0.938          & \multicolumn{1}{c|}{0.945}          & \multicolumn{1}{c|}{0.907}          & \multicolumn{1}{c|}{0.925}          & 0.958          \\ \hline
HRC$^\#$ & \multicolumn{1}{c|}{0.941}          & \multicolumn{1}{c|}{0.905}          & \multicolumn{1}{c|}{0.935}          & 0.949          & \multicolumn{1}{c|}{0.941}          & \multicolumn{1}{c|}{0.909}          & \multicolumn{1}{c|}{0.933}          & 0.942          & \multicolumn{1}{c|}{0.947}          & \multicolumn{1}{c|}{0.906}          & \multicolumn{1}{c|}{0.927}          & 0.961          \\ \hline
HRC$^!$      & \multicolumn{1}{c|}{0.941}          & \multicolumn{1}{c|}{0.907}          & \multicolumn{1}{c|}{0.934}          & 0.951          & \multicolumn{1}{c|}{0.943}          & \multicolumn{1}{c|}{0.906}          & \multicolumn{1}{c|}{0.931}          & 0.937          & \multicolumn{1}{c|}{0.951}          & \multicolumn{1}{c|}{0.911}          & \multicolumn{1}{c|}{0.929}          & 0.961          \\ \hline
\textbf{HRC}         & \multicolumn{1}{c|}{\textbf{0.943}} & \multicolumn{1}{c|}{\textbf{0.913}} & \multicolumn{1}{c|}{\textbf{0.938}} & \textbf{0.959} & \multicolumn{1}{c|}{\textbf{0.948}} & \multicolumn{1}{c|}{\textbf{0.915}} & \multicolumn{1}{c|}{\textbf{0.937}} & \textbf{0.953} & \multicolumn{1}{c|}{\textbf{0.955}} & \multicolumn{1}{c|}{\textbf{0.919}} & \multicolumn{1}{c|}{\textbf{0.935}} & \textbf{0.969} \\ \hline
\end{tabular}
\label{ablation_study}
\vspace{-0.3cm}
\end{table*}}
\end{center}

\vspace{-1.5cm}
\subsubsection{Evaluation Metrics}
To show the effectiveness of the proposed method, we use the accuracy metric for evaluating the classification of machine learning tasks. Additionally, we use the following metrics: 

\begin{itemize}
    \item {\textbf{Accuracy} is the ratio of the sum of True Positives and True Negatives to the sum of True Positives, True Negatives, False Positives, and False Negatives. Formally, the accuracy is given by $\frac{TP+TN}{TP+TN+FP+FN}$, where $TP$, $TN$, $FP$ and $FN$ represent the number of True Positives, True Negatives, False Positives and False Negatives respectively. Larger values indicate better performance.}
    \item { \textbf{Precision} is given by$\frac{TP}{TP+FP}$ and presents the ratio of the number of True Positives to the number of True Positives and False Positives. Larger values indicate better results.}
    \item {\textbf{Recall} is given by $\frac{TP}{TP+FN}$ and presents the ratio of the number of True Positives to the number of True Positives and False Negatives. Larger values indicate better results.}
    \item {\textbf{F-Measure} considers both precision and recall in a single metric by taking their harmonic mean. Formally, F-measure is given by $2*P*R/(P+R)$, where $P$ and $R$ are precision and recall, respectively. Larger values indicate better results.}
\end{itemize}

\subsubsection{Baseline Methods}
We compare the performances of our proposed methods: hierarchical reinforcement crossing feature generation (HRC) against the following four baseline algorithms.

\begin{itemize}
    \item {\textbf {Raw:} Directly conduct machine learning algorithms on the original dataset without any feature generation.}
    \item {\textbf {DeepFM:} DeepFM~\cite{guo2017deepfm} combines factorization machines and deep learning in a new neural network architecture. It derives an end-to-end feature generation model that emphasizes both low and high-order feature interactions.}
    \item {\textbf {xDeepFM:} xDeepFM~\cite{lian2018xdeepfm} generates feature interactions in an explicit fashion and at the vector-wise level. It can also learn arbitrary low and high-order feature interactions implicitly.}
    \item {\textbf {AutoCross:} AutoCross~\cite{luo2019autocross} is the state-of-the-art method that outperforms other existing feature generation methods based on cross operation. It can automatically construct high-order cross-features and uses a beam search strategy to iteratively generate a locally optimal feature set.}
    \item {\textbf{GRFG}~\cite{wang2022group} studies the problem of generation on continuous features and generates features based on operations (e.g.,  +, -, *, sin, cos).}
\end{itemize}

Besides, we also implement ablation studies to verify the effectiveness of our method. 

\begin{itemize}
    \item {\textbf{HRC$^*$:} which means that we replace the controller with a greedy strategy based on our hierarchical strategy.}
    \item { \textbf{HRC$^\#$:} which means that we replace the meta controller with a greedy strategy based on our hierarchical strategy.}
    \item {\textbf{HRC$^!$:} which represents that we don't take any reinforcement learning method.}
\end{itemize}

\subsubsection{Hyperparameters and Reproducibility}
In the experiments, for all HRC, we set the batch size to 20 and used AdamOptimizer with a learning rate of 0.01. We set the $Q$ network of meta controller and controller in our methods as a one-layer ReLU with 100 nodes, respectively. The memory is set to 40 for each DQN. We limited the episode to 15, with each episode consisting of 70 exploration steps. All datasets are split with 80\% for training and 20\% for testing. The weights in reward functions have equal values. We used logistic regression as the default downstream task considering the fact that logistic regression is the most popular algorithm that is in need of feature generation in real-world applications~\cite{luo2019autocross}.

\begin{figure*}[hbtp]
	\centering
	\subfigure[Access]{\label{exp:access}\includegraphics[width=4.35cm]{{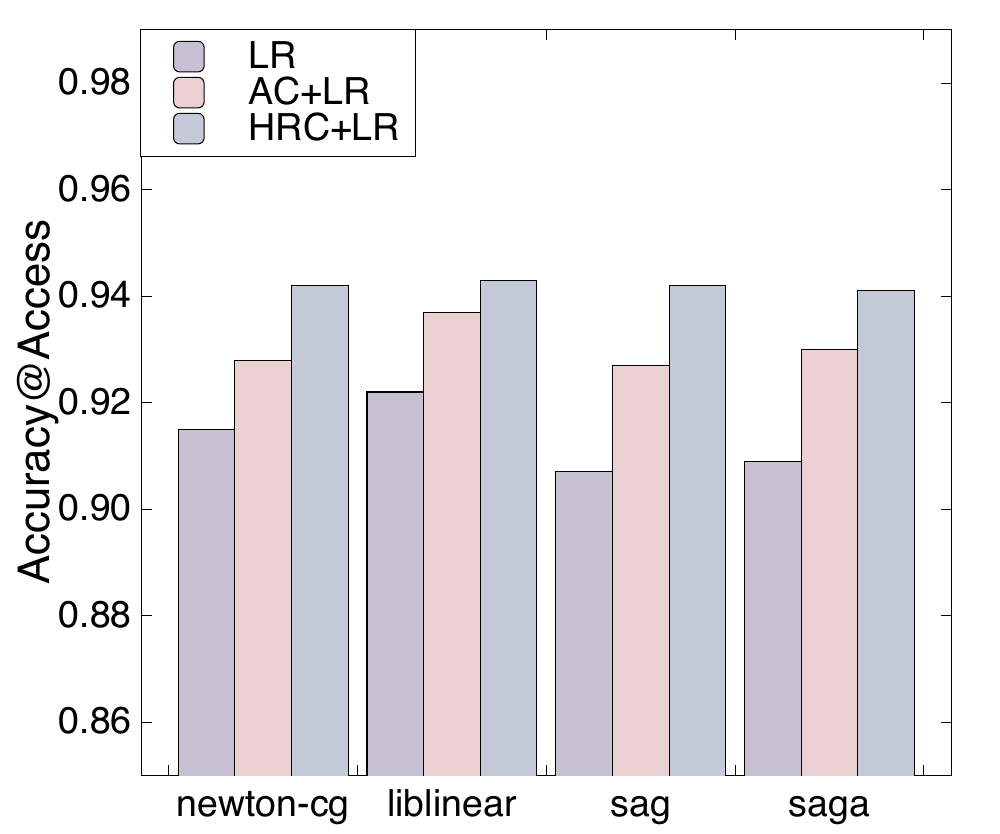}}}
	\subfigure[Bank]{\label{exp:bank}\includegraphics[width=4.35cm]{{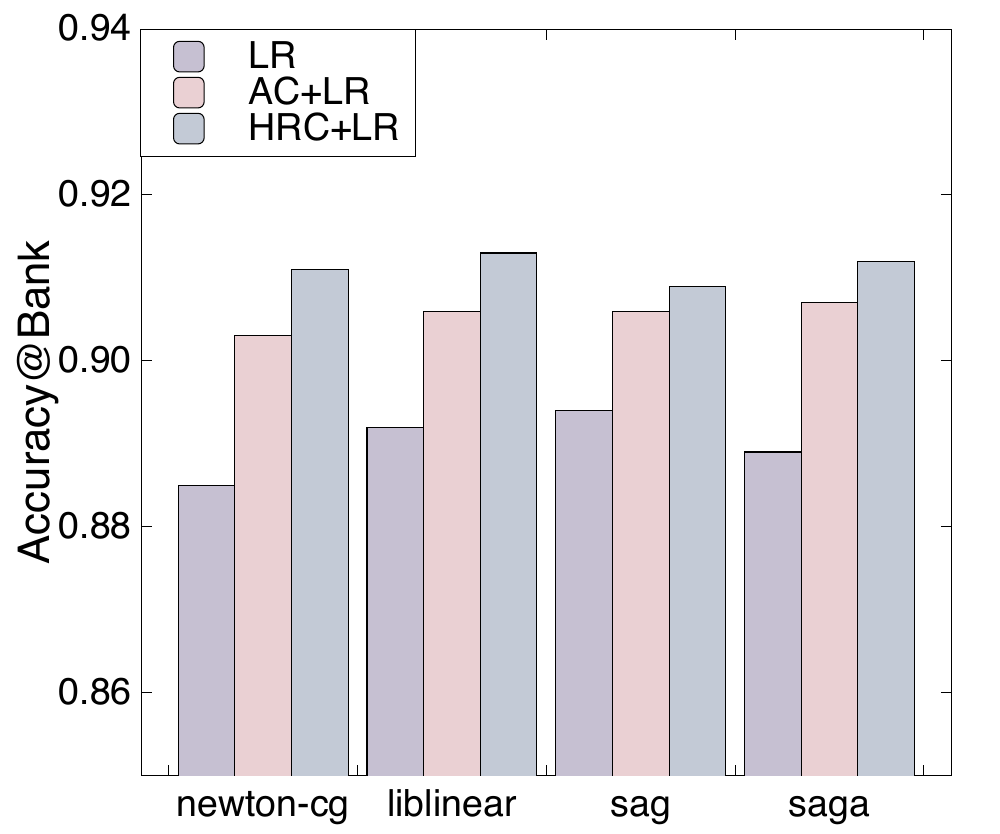}}}
	\subfigure[Credit]{\label{exp:credit}\includegraphics[width=4.35cm]{{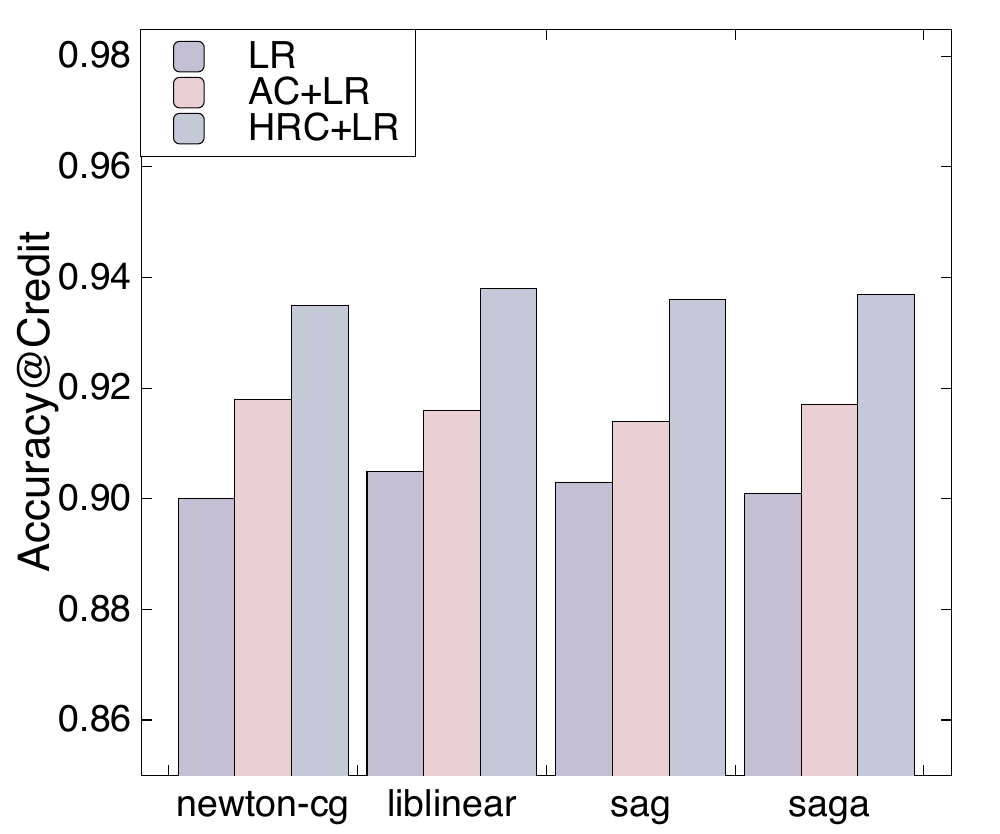}}}
	\subfigure[Nomao]{\label{exp:nomao}\includegraphics[width=4.35cm]{{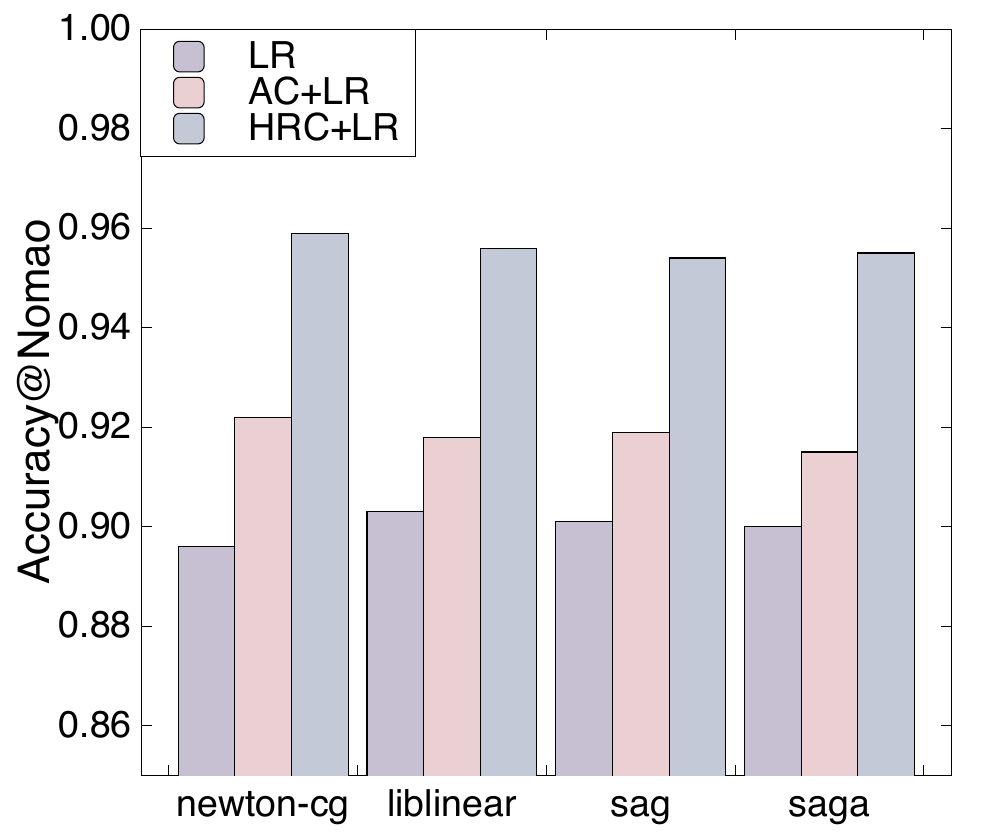}}}
	\caption{Performance comparison of different solvers on different datasets.}
	\label{fig:robustness}
 \vspace{-0.5cm}
\end{figure*}

\begin{figure*}[hbtp]
	\centering
	\subfigure[Accuracy]{\label{exp:Accuracy}\includegraphics[width=4.35cm]{{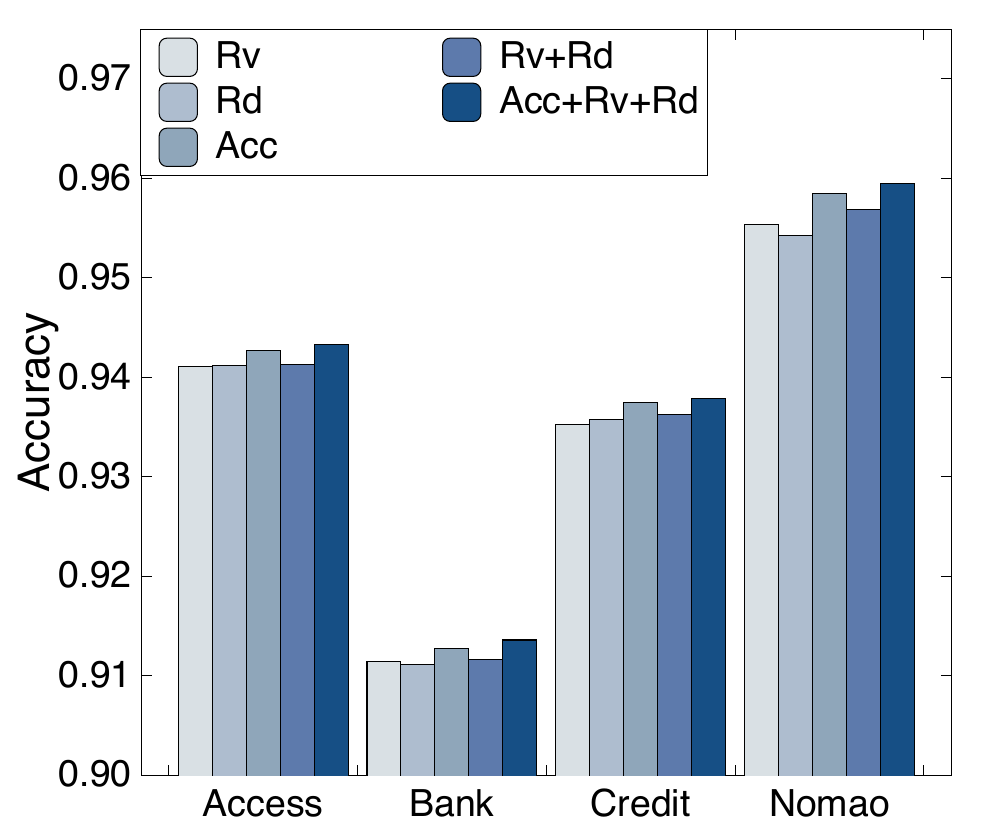}}}
	\subfigure[Precision]{\label{exp:Precision}\includegraphics[width=4.35cm]{{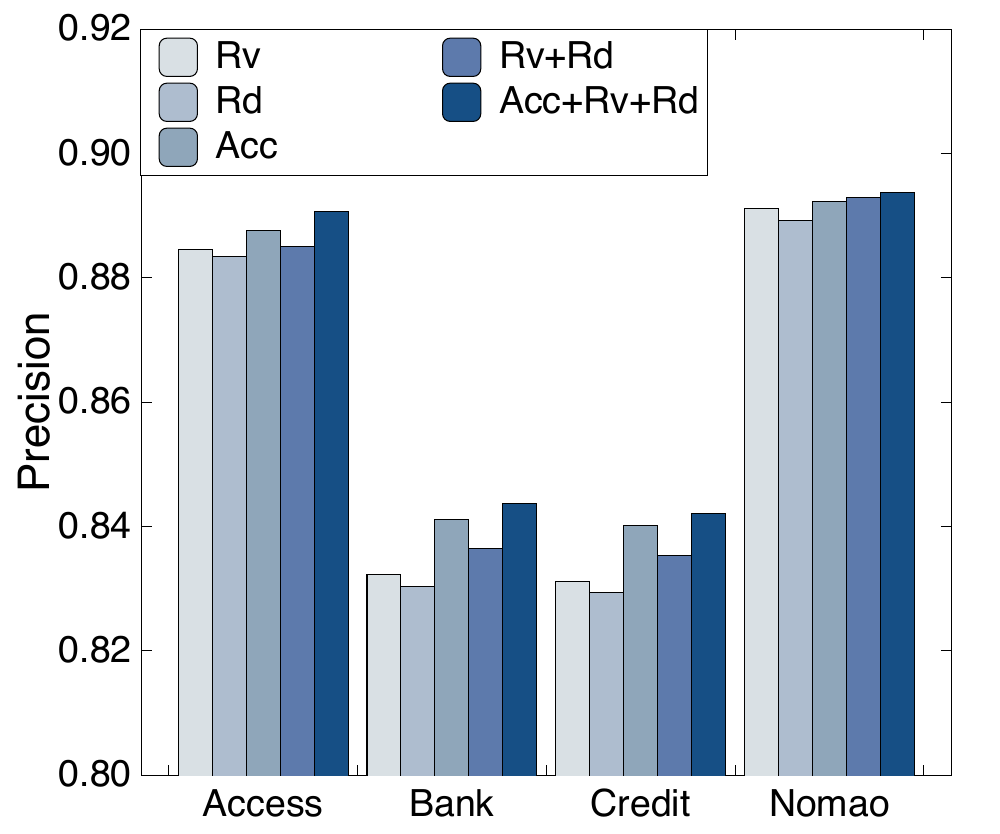}}}
	\subfigure[Recall]{\label{exp:Recall}\includegraphics[width=4.35cm]{{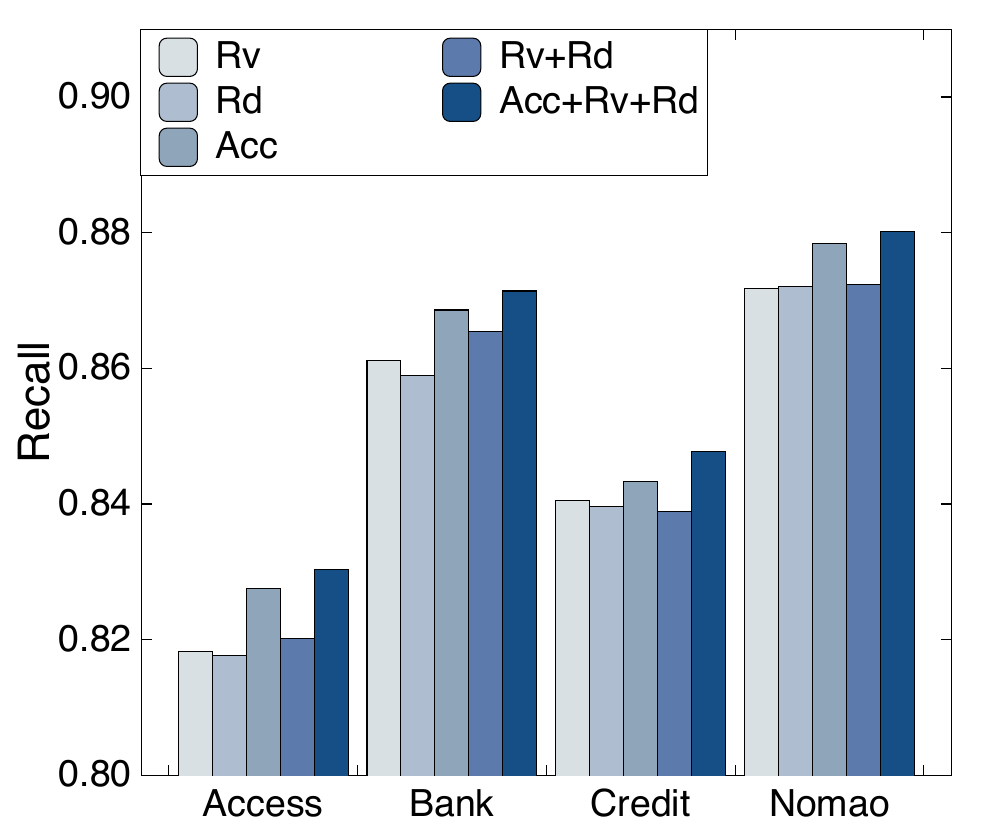}}}
	\subfigure[F-Measure]{\label{exp:F-Measure}\includegraphics[width=4.35cm]{{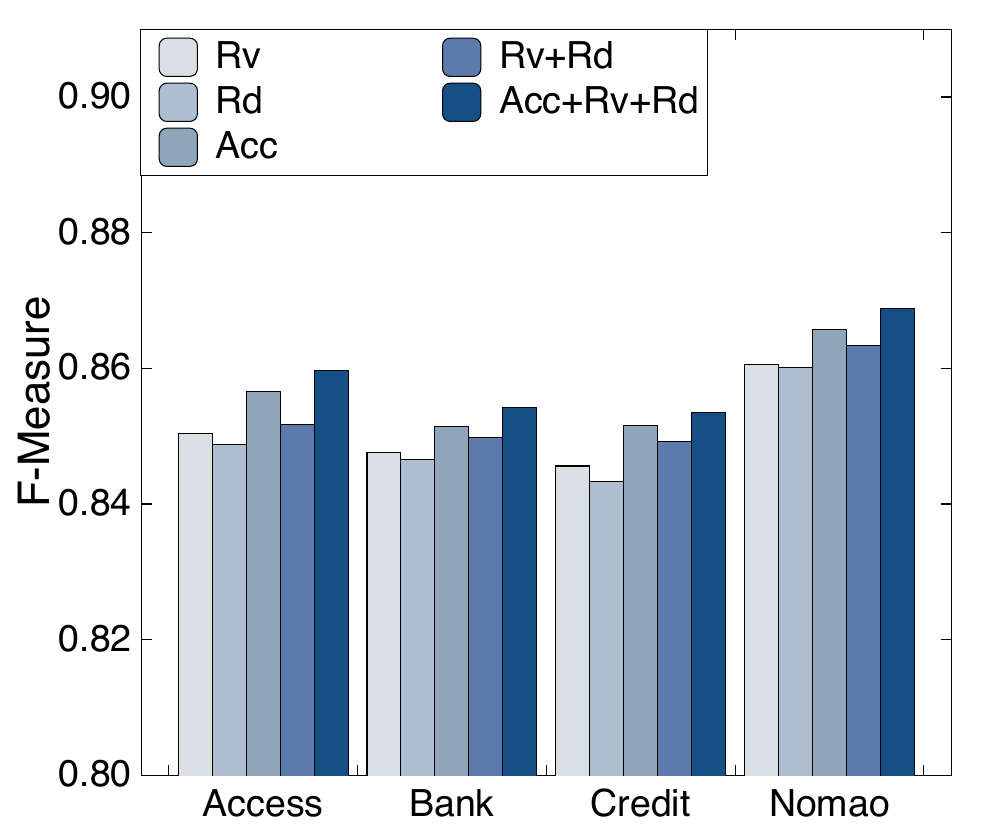}}}
	\caption{Performance comparison of different rewards in HRC on different datasets.}
	\label{fig:reward}
 \vspace{-0.5cm}
\end{figure*}
\vspace{-0.1cm}
\subsection{Experiment Results}

\subsubsection{Overall Performance}
This experiment aims to answer: \emph{Whether our method can generate the best cross-feature space and improve downstream tasks}. Here, we would first like to emphasize that the classification task is the focus of all of our experiments. We take the approach of discretizing continuous values for feature crossing. However, in the regression task, this method will lose data accuracy when discretizing continuous variables. And we discovered that the regression task just got minor improvement during our experimental investigation. \textbf{Table \ref{overall_performance}} shows the experimental results on different classification tasks in terms of accuracy, respectively. We observe that HRC always yields the best results compared to the available solutions for different datasets and different downstream tasks. The potential reason is that the HRC approach uses a two-stage reinforcement learning task in feature-cross that takes into account the overall optimization reward of long-term steps, allowing the model to search for the globally optimal result quickly. In addition, we do not need to enumerate all combinations of feature-cross, so it is very fast and efficient to get the best performance. Thus, compared with state-of-the-art baselines, our method is more practical and accurate in real-world application scenarios. 

\subsubsection{Ablation Study}
This experiment aims to answer: \emph{The impact of reinforcement learning component on the final result.} In our method, we design two DQNs for the meta controller and the controller, respectively. So in this set of experiments, we replace any of the reinforcement learning components with a greedy algorithm to verify the effectiveness of the HRC. \textbf{Table \ref{ablation_study}} shows the results of ablation studies on different tasks evaluated by accuracy. We can discover that the results of HRC$^*$ are the worst in most cases. The HRC$^\#$ and HRC$^!$ perform slightly better than the HRC$^*$ but still perform worse than the HRC. Therefore, these experiments indicate that taking account of long-term rewards is indeed an important approach to improving feature generation.

\subsubsection{Robustness Check}
This experiment aims to answer: \emph{Whether HRC can still obtain stable results for different downstream tasks}. The predictive performance relies on not just feature generation, but also downstream tasks. As a result, we apply our method to logistic regression with different kernels in the generated feature set to see if our generated feature set is consistently stable and can consistently outperform other baseline methods on various predictor settings. In this way, we can examine the robustness of our methods. We use
(1) `newton-cg' solver; (2) `liblinear' solver; (3) `sag' solver; (4) `saga' solver as the solvers of logistic regression for this experiment. We compare our method with LR (short for logistic regression) and the state-of-the-art AC (short for AutoCross) +LR. \textbf{Figure \ref{fig:robustness}} shows the comparison of different solvers on different datasets. We can see that in every solver, The HRC+LR method outperforms AC+LR.

\subsubsection{Study of Reward in HRC}
This experiment aims to answer: \emph{The impact of reward function design on the final result}. We study the impacts of the reward function in HRC and consider five cases: 

(1) \textbf{Rv} that only considers relevance in the reward function; 

(2) \textbf{Rd} that only considers redundancy in the reward function; 

(3) \textbf{Acc} that only considers accuracy in the reward function; 

(4) \textbf{Rv+Rd} that only considers relevance and redundancy in the reward function;

(5) \textbf{Acc+Rv+Rd} that considers accuracy, relevance and redundancy in the reward function.

\begin{figure*}[]
	\centering
	\subfigure[Access]{\label{exp:access}\includegraphics[width=4.3cm]{{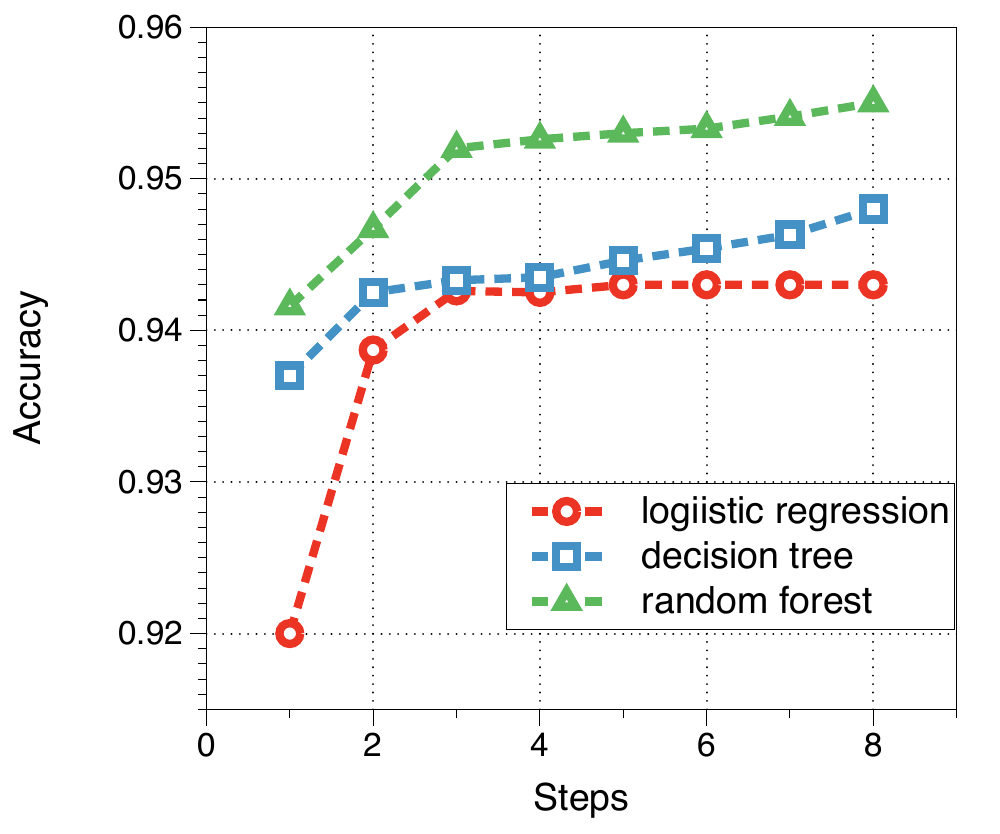}}}
	\subfigure[Bank]{\label{exp:bank}\includegraphics[width=4.3cm]{{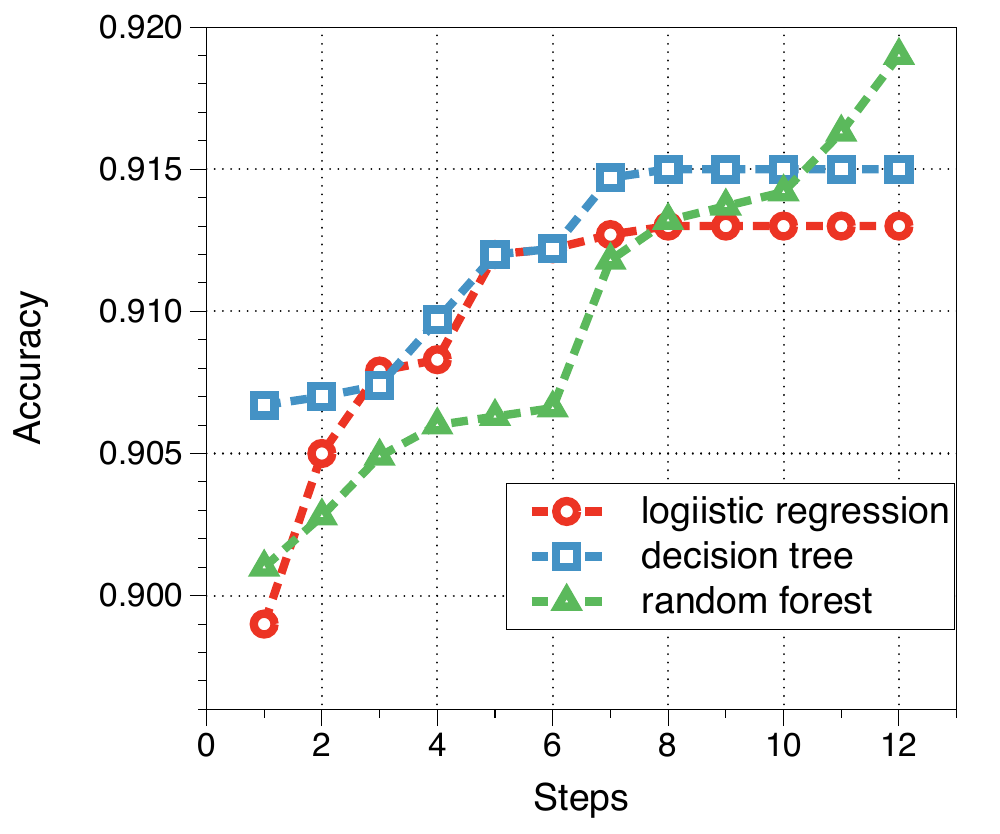}}}
	\subfigure[Credit]{\label{exp:credit}\includegraphics[width=4.3cm]{{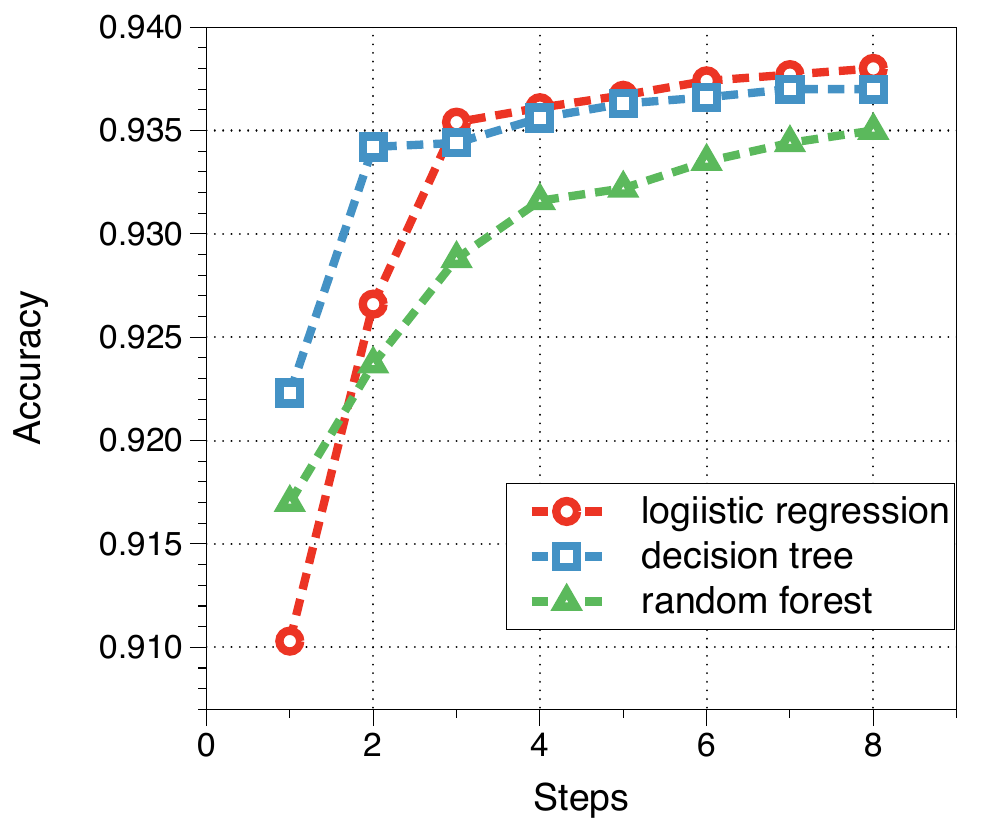}}}
	\subfigure[Nomao]{\label{exp:nomao}\includegraphics[width=4.3cm]{{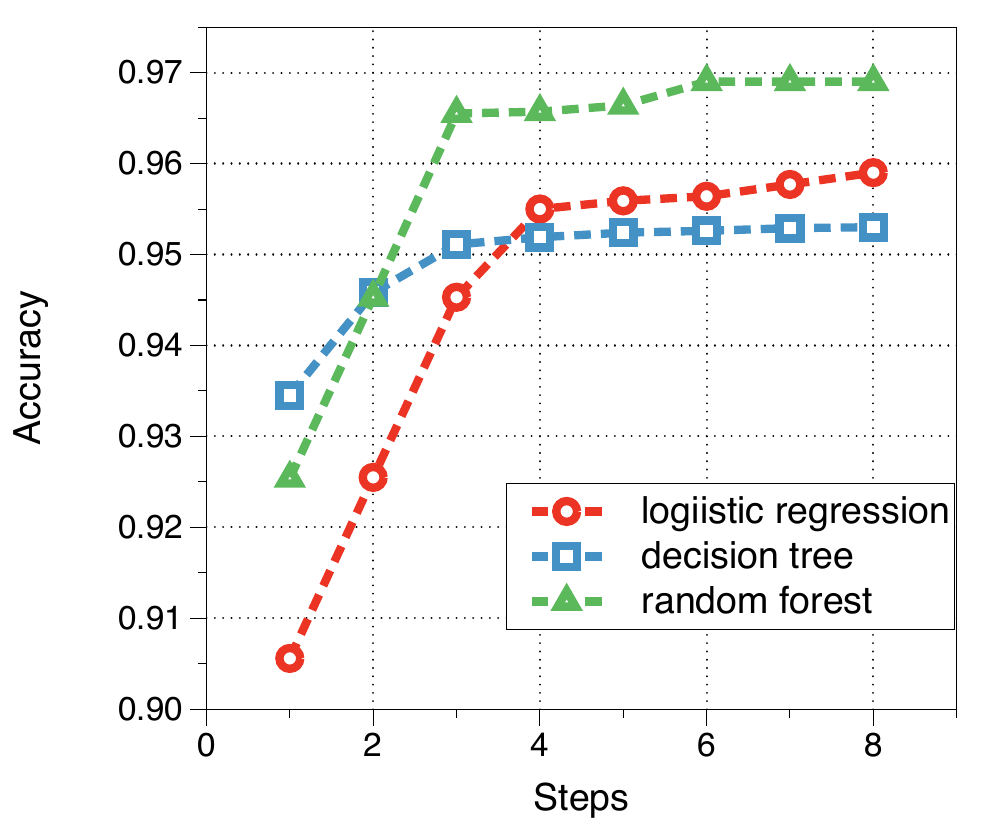}}}
	\caption{Convergence steps of different downstream tasks on different datasets.}
	\label{fig:converge}
 \vspace{-0.5cm}
\end{figure*}
\textbf{Figure \ref{fig:reward}} reveals the performance comparison of different rewards in HRC on different datasets. We can see that Acc is the second-best reward function, since it leads exploration in the direction of improving accuracy. Rv and Rd are less satisfactory. This is because both are unsupervised indicators of rewards and are not directly relevant to prediction accuracy. Their combination of Rv and Rd improves performance slightly but does not outperform Acc. Acc+Rv+Rd achieves the best performance since it takes both supervised and unsupervised indicators into account.

\subsubsection{Convergence of HRC}
This experiment aims to answer: \emph{How many steps would the HRC take before convergent}? It is important for reinforcement learning methods to converge. We study the performances of HRC with different episodes, varying from 0 to 15 episodes over different datasets.  Normally, the convergence of reinforcement learning methods is evaluated by the steadiness of accumulated rewards, but this is not the case in the feature generation problem. In feature generation, we only need to remember one optimal feature set through all of the episodes, and the HRC feature generation method is considered to converge when the remembered feature set no longer changes or falls into a predefined range. \textbf{Figure \ref{fig:converge}} shows that our HRC feature generation method converges in several episodes, which is extremely fast for a reinforcement learning method.

%% file: related_work.tex
\vspace{-0.3cm}
\section{Related Work}\label{related_work}
\vspace{-0.1cm}
\noindent \textbf{Automated Feature Generation.} There are three major categories of automated feature generation methods, i.e., factorization machine-based methods, cross-operation-based methods, and embedded methods~\cite{xiaomeng1, chandrashekar2014survey, saeys2007review, xiao2023traceable}. Factorization machine methods can effectively capture the low-order interactions between features. {\it Blondel et al.} proposed the first generic yet efficient algorithms for training arbitrary-order HOFMs~\cite{blondel2016higher}. {\it Cheng et al.} proposed a novel Gradient Boosting Factorization Machine (GBFM) model to incorporate a feature selection algorithm with Factorization Machines into a unified framework~\cite{cheng2014gradient}. {\it Juan et al.} established ﬁeld-aware factorization machines as an eﬀective method for classifying large sparse data including those from CTR prediction~\cite{juan2016field}. {\it Cheng et al.} presented Wide and Deep learning-jointly trained wide linear models and deep neural networks to combine the beneﬁts of memorization and generalization for recommender systems~\cite{cheng2016wide}. These methods were implicit and had difficulty to capture high-order interactions. Cross operations such as~\cite{luo2019autocross,shan2016deep} captured high-order feature interactions. {\it Liu et al.} gave a definition of interpretation inconsistency in deep neural networks, and proposed a novel method called CrossGO, which selected useful cross features according to the interpretation inconsistency~\cite{liu2019automatically}. {\it Luo et al.} proposed successive mini-batch gradient descent and multi-granularity discretization to further improve efficiency and effectiveness, while ensuring simplicity so that no machine learning expertise or tedious hyper-parameter tuning was required~\cite{luo2019autocross}. There were also other embedded methods like gradient boost machine~\cite{friedman2001greedy} and group lasso~\cite{meier2008group} which build useful features in the process of model training. {\it Dong et al.} combined feature selection and feature generation into a transformation graph and optimized the two processes jointly~\cite{dong2018feature}. However, these methods usually require complex optimization procedures and thus have difficulties dealing with large-scale datasets~\cite{autogeneration1, autogeneration2}. 

\noindent \textbf{Hierarchical Reinforcement Learning.}
Hierarchical reinforcement Learning (HRL)~\cite{HRL1} decomposes the target Markov decision process (MDP)~\cite{PUTERMAN1990331}into a hierarchy of smaller MDPs and solves them sequentially~\cite{dietterich2000hierarchical,botvinick2012hierarchical,barto2003recent}. There are numerous studies on HRL. {\it Dietterich et al.} defined a hierarchical Q learning algorithm, proved its convergence, and showed experimentally that it can learn much faster than ordinary “flat” Q learning~\cite{dietterich1998maxq}. {\it Vezhnevets et al.} employed a Manager module and a Worker module. The Manager operated at a lower temporal resolution and set abstract goals which are conveyed to and enacted by the Worker. The Worker generated primitive actions at every tick of the environment~\cite{vezhnevets2017feudal}. {\it Morimoto et al.} proposed a hierarchical reinforcement learning architecture that realized practical learning speed in real hardware control tasks~\cite{morimoto2001acquisition}. {\it Ribas et al.} proposed that the computations supporting hierarchical behavior may relate to those in HRL, a machine-learning framework that extended reinforcement-learning mechanisms into hierarchical domains~\cite{ribas2011neural}. {\it Florensa et al.} proposed a general framework that first learns useful skills in a pre-training environment, and then leverages the acquired skills for learning faster in downstream tasks~\cite{florensa2017stochastic}. {\it Lin et al.} proposed to tackle the large-scale fleet management problem using reinforcement learning, and proposed a contextual multi-agent reinforcement learning framework that successfully tackled the taxi fleet management problem~\cite{lin2018efficient}. {\it Wei et al.} proposed a more effective deep reinforcement learning model for traffic light control~\cite{wei2018intellilight}. {\it Nachum et al.} proposed to use off-policy experience for both higher and lower-level training~\cite{nachum2018data}.

%% file: conclusion.tex
\vspace{-0.2cm}
\section{Conclusion}
\vspace{-0.1cm}
In this paper, we study the problem of automated feature generation. We propose the hierarchical reinforcement crossing (HRC) method to generate the cross-feature set, which can obtain the best performance on all datasets. In the HRC method, we fuse predictive accuracy and information gain to design an inspiring reward function; we use the categorical hashing representation to reduce the dimension of an increased feature set; and we design a descriptive state representation method to derive a fixed-length state vector as the input for reinforcement learning policies. The challenges of the feature generation problem are that the number of generated features can be explosively large and the generated feature set extremely falls into local optima. Our method doesn't need to enumerate all potential combinations of features, and it can explore a globally optimal cross-feature set quickly considering the long-term rewards. The performance superiority in experiments demonstrates the effectiveness of our approach.